\documentclass[letterpaper, 10 pt, conference]{template/ieeeconf}

\IEEEoverridecommandlockouts                              

\usepackage{graphicx}
\usepackage{caption}
\usepackage{amsmath,amssymb,enumerate}

\usepackage{amsthm}
\usepackage{mathtools}
\usepackage{breqn}
\usepackage{algorithm, algorithmicx, algpseudocode}

\usepackage{blindtext}
\usepackage{gensymb}
\usepackage{xparse}
\usepackage{lipsum}
\usepackage{mathrsfs}
\usepackage[mathscr]{euscript}
\usepackage{times}
\usepackage{cite} 
\usepackage{multicol}
\usepackage[caption=false,font=footnotesize]{subfig}
\usepackage{amsfonts}
\usepackage[utf8]{inputenc}
\usepackage[T1]{fontenc}
\usepackage{textcomp}
\usepackage{amsfonts}
\usepackage{soul}
\usepackage{multirow}
\usepackage{booktabs}
\usepackage[usenames,dvipsnames,svgnames,table, xcdraw]{xcolor}

\usepackage[colorlinks=true,pdfpagemode=UseNone,citecolor=black,linkcolor=black,urlcolor=BrickRed]{hyperref}

\usepackage{comment}

\newtheorem{remark}{Remark}

\renewcommand{\thefigure}{\arabic{figure}}

\newcommand{\squeezeup}{\vspace{-3mm}}

\newcommand{\m}{\mathop{\mathrm{m}}}

\newcommand{\transpose}{\mathsf{T}}

\DeclareDocumentCommand{\vectorToSkew}{ O{} }{\left(#1\right)_\times}

\usepackage{rotating}
\usepackage{outlines}
\usepackage{multirow}
\usepackage{caption}

\title{\LARGE \bf
Monocular Depth Prediction through Continuous 3D Loss
}

\author{Minghan Zhu$^{1}$, Maani Ghaffari$^{1}$, Yuanxin Zhong$^{1}$, Pingping Lu$^{1}$, \\ Zhong Cao$^{2}$, Ryan M. Eustice$^{1}$ and Huei Peng$^{1}$
\thanks{*This work was partially supported by the Toyota Research Institute (TRI), partly under award number N021515.}%
\thanks{$^{1}$M.~Zhu, M.~Ghaffari, Y.~Zhong, P.~Lu, R.~Eustice and H.~Peng are with the University of Michigan, Ann Arbor, MI 48109, USA. {\tt\small\{minghanz, maanigj, zyxin, pingpinl, eustice,  hpeng\}@umich.edu}}%
\thanks{$^{2}$Z.~Cao is with Tsinghua University, Beijing, 100084, China. {\tt\small caoc15@mails.tsinghua.edu.cn}}%
}

\newcommand{\cell}[1]{\begin{tabular}{@{}l@{}}#1\end{tabular}}

\begin{document}

\maketitle
\thispagestyle{empty}
\pagestyle{empty}

\begin{abstract}

This paper reports a new continuous 3D loss function for learning depth from monocular images. The dense depth prediction from a monocular image is supervised using sparse LIDAR points, which enables us to leverage available open source datasets with camera-LIDAR sensor suites during training. Currently, accurate and affordable range sensor is not readily available. Stereo cameras and LIDARs measure depth either inaccurately or sparsely/costly. In contrast to the current point-to-point loss evaluation approach, the proposed 3D loss treats point clouds as continuous objects; therefore, it compensates for the lack of dense ground truth depth due to LIDAR's sparsity measurements. We applied the proposed loss in three state-of-the-art monocular depth prediction approaches DORN, BTS, and Monodepth2. Experimental evaluation shows that the proposed loss improves the depth prediction accuracy and produces point-clouds with more consistent 3D geometric structures compared with all tested baselines, implying the benefit of the proposed loss on general depth prediction networks. A video demo of this work is available at \url{https://youtu.be/5HL8BjSAY4Y}.
\end{abstract}

\section{INTRODUCTION}


Range measurement is vital for robots and autonomous vehicles. For ground vehicles, reliable and accurate range sensing is the key for Adaptive Cruise Control, Automatic Emergency Braking, and autonomous driving. With rapid development in deep learning techniques, image-based depth prediction gained much attention and progress, promising cost-effective and accessible range sensing using commercial monocular cameras. However, depth ground truth for an image is not always available for training a neural network. Today, in outdoor scenarios, we mainly rely on LIDAR sensors to provide accurate and detailed depth measurements, but the point clouds are too sparse compared with image pixels. Besides, LIDARs cannot get reliable reflection on some surfaces (e.g. dark, reflective, transparent \cite{140_eigen2014depth}). Using stereo cameras is another way for range sensing, but it is less accurate for mid to far distance. Generating ground truth depth from an external visual SLAM module \cite{167_yang2018deep,181_wang2018learning} suffers similar problems, subject to noise and error. 


\begin{figure}[t]
\begin{tabular*}{\columnwidth}{l@{\extracolsep{\fill}}r}
  \cell{\includegraphics[width=0.9\columnwidth]{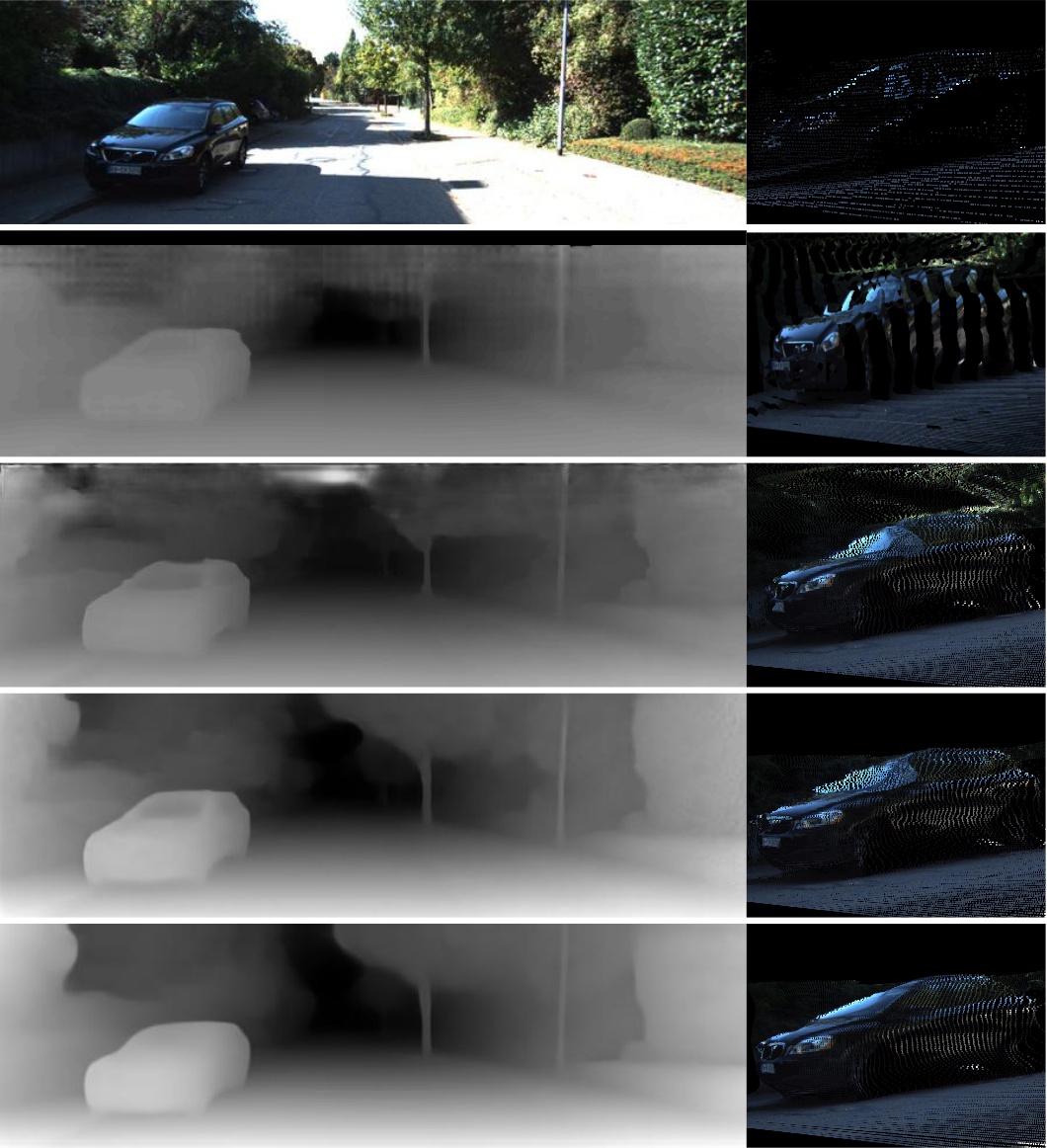}} & 
  \cell{{\rotatebox[origin=c]{90}{\parbox[c]{30pt}{\centering\footnotesize Image \& LIDAR}}}    \\ 
        {} \\
        {\rotatebox{90}{\footnotesize DORN\cite{dorn_fu2018deep}}}  \\ 
        {} \\
        {\rotatebox{90}{\footnotesize BTS\cite{bts_lee2019big}}}  \\ %
        {} \\
        {\rotatebox{90}{\centering\footnotesize Monodepth2\cite{155_godard2019digging}}}  \\ %
        {} \\
        {\rotatebox{90}{\footnotesize Ours}}}
\end{tabular*}
\caption{Visualization of depth prediction. \textbf{1st row:} image and raw LIDAR scan of the vehicle colored by the image. \textbf{2nd - 4th rows:} depth predictions and point-clouds generated from image pixels with predicted depth using baseline methods. \textbf{5th row:} Our results. Our method can build on general depth prediction networks. We tested our method on the above three networks, but the figure only shows our result based on Monodepth2 \cite{155_godard2019digging} network for simplicity. This data sample is from KITTI dataset \cite{KITTIRaw_Geiger2013IJRR}. }
    \label{fig:firstpage}
\end{figure}
Due to the lack of perfect ground truth, as discussed above, and the fact that monocular cameras are prevalent, much research effort has been devoted to unsupervised monocular depth learning, which requires only sequences of monocular images as training data. These approaches have shown promising progress, but there is still a performance gap to supervised approaches (see Table~\ref{tab:kitti}). Moreover, monocular unsupervised approaches are inherently scale-ambiguous. The depth prediction is relative and needs a scale factor to recover the true depth, meaning that there are real deployment limitations. 


Despite that LIDAR sensors are still too expensive for large-scale deployment on vehicles, a number of driving datasets with sensor suites including cameras and LIDARs are already available \cite{waymo_open_dataset,lyft2019,nuscenes2019,dong2019mcity}. Given such rich datasets, we improve monocular depth prediction by leveraging sparse LIDAR data as ground truth. As stated in BTS\cite{bts_lee2019big}, currently ranking 1st in monocular depth prediction using the KITTI dataset \cite{KITTIRaw_Geiger2013IJRR} (Eigen's split \cite{140_eigen2014depth}), the high sparsity of ground truth data limits the depth prediction accuracy. Addressing the same issue, we propose a new continuous 3D loss that transforms discrete point clouds into continuous functions. The proposed loss better exploits data correlation in Euclidean and feature spaces, leading to improved performance of the current deep neural networks. An example is shown in Fig.~\ref{fig:firstpage}. We note that the proposed 3D loss function is agnostic to the network architecture design, an active research area.
The main contributions of this paper include:
\begin{enumerate}
    \item We propose a novel continuous 3D loss function for monocular depth prediction.
    \item By merely adding this loss to several state-of-the-art monocular depth prediction approaches~\cite{155_godard2019digging, dorn_fu2018deep, bts_lee2019big}, without modifying the network structures, we obtain more accurate and geometrically-plausible depth predictions compared with all these baseline methods on KITTI dataset under the supervision of raw LIDAR points. 
    \item Our work is open-sourced and software is available for download at \url{https://github.com/minghanz/c3d}. 
\end{enumerate}

The remainder of this paper is organized as follows. The literature review is given in Sec. \ref{sec:review}. The proposed new loss function, the theoretical foundation, and its application in monocular depth prediction are introduced in Sec. \ref{sec:method}. The experimental setup and results are presented in Sec. \ref{sec:experiments}. Section~\ref{sec:conclusions} concludes the paper and provide future work ideas.

\section{Related Work} \label{sec:review}



Deep-learning-based 3D geometric understanding shares similar ideas with SfM/vSLAM approaches. For example, the application of reprojection loss in unsupervised depth prediction approaches \cite{174_zhou2017unsupervised} and direct methods in SfM/vSLAM \cite{newcombe2011dtam} are tightly connected. However, they are fundamentally different since the back-propagation of neural networks only takes a small step along the gradient to learn the general prior from large amounts of data gradually. Learning correspondences among different views can assist with recovering the depth \cite{143_ummenhofer2017demon} if stereo or multi-view images are available as input. For single-view depth prediction, the network needs to learn from more general cues, including perspective, object size, and scenario layout. Although single-view depth prediction is an ill-posed problem in theory since infinite possibilities of 3D layout could result in the same 2D rendered image, this task is still viable since the plausible geometric layouts occur in the real world is limited and can be learned from data. 

\subsection{Supervised single-view depth prediction}
It is straight-forward to learn image depth by minimizing the point-wise difference between the predicted depth value and the ground truth depth value. The ground truth depth can come from LIDAR, but such measurements are sparse. One strategy is simply masking out pixels without ground truth depth values and only evaluating loss on valid points~\cite{140_eigen2014depth}. An alternative is to fill in invalid pixels in ground truth maps before evaluation~\cite{141_laina2016deeper}, for example using ``colorization'' methods~\cite{levin2004colorization} included in NYU-v2 dataset~\cite{NYUv2_silberman2012indoor}. While learning from the preprocessed dense depth maps is an easier task, it also limits the accuracy upper bound. The work of \cite{ICCV1911_bozorgtabar2019syndemo,ICCV1912_di2019monocular} used synthetic datasets (e.g. \cite{VKITTI_gaidon2016virtual,ros2016synthia}) for training, in which perfect dense ground truth depth maps are available. However, in practice, the domain difference between synthetic and real data poses a challenge.

\subsection{Unsupervised single-view depth prediction}
The fact that an image's ground truth depth is hard to obtain and usually sparse and noisy motivates some researchers to apply unsupervised approaches. Stereo cameras with known baseline provide self-supervision in that an image can be reconstructed from its stereo counterpart if the disparity is accurately estimated. Following this idea \cite{144_garg2016unsupervised} proposed an end-to-end method to learn single-view depth from stereo data. Using consecutive image frames for self-supervision is similar, except that the camera motions between the consecutive time steps must be estimated and that scale ambiguity may arise. The work of \cite{174_zhou2017unsupervised} is one of the first proposing to use monocular videos only to learn pose and depth prediction through CNNs in an end-to-end manner. Researchers included an optical flow estimation module \cite{148_yin2018geonet} and a motion segmentation module to deal with moving objects \cite{156_ranjan2019competitive} so that rigid and non-rigid parts are treated separately. 

\subsection{Loss functions in single-view depth prediction}
Existing learning methods mainly rely on direct supervision of true depth and indirect supervision of view synthesis error. Most other loss functions are regularization terms. We summarize commonly used loss functions in the following. We omit loss functions from the adversarial learning framework~\cite{159_aleotti2018generative}, as they require dedicated network structures. 

\subsubsection{Geometric losses} 
Point-wise difference between predicted and ground truth depth values in the norms of L1 \cite{ICCV1920_wu2019spatial}, L2 \cite{147_qi2018geonet}, Huber \cite{CVPR1904_ye2019student}, berHu \cite{141_laina2016deeper}, and the same norms of inverse depth \cite{167_yang2018deep} have all been applied, with the consideration of emphasizing prediction error of near/far points. Cross-entropy loss \cite{ICCV1904_Shin_2019_ICCV} and ordinal loss \cite{180_fu2018deep} are applied when depth prediction is formulated as a classification or ordinal regression, instead of regression problems. The negative log-likelihood is adopted in approaches producing probabilistic outputs, e.g., in \cite{ICCV1906_Brickwedde_2019_ICCV}. \cite{140_eigen2014depth} introduced a scale-invariant loss to enable learning from data across scenarios with large scale variance. The surface normal difference is also a form of more structured geometric loss \cite{147_qi2018geonet}. In contrast to the above loss terms which takes value difference in the image space, \cite{154_mahjourian2018unsupervised} directly measure geometric loss in the 3D space, minimizing point cloud distance by applying ICP (Iterative Closest Point) algorithms. \cite{ICCV1914_Yin_2019_ICCV} proposed non-local geometric losses to capture large scale structures. 

\subsubsection{Non-geometric loss}
This class of loss functions is applied in unsupervised approaches. The most commonly used forms are intensity difference between warped and original pixels, and Structured Similarity (SSIM) \cite{SSIM}, which also captures the higher-order statistics of pixels in a local area. In order to handle occlusion and non-rigid scenarios, various adjustments to the photometric errors are proposed. For example, using weight or masking to ignore a subset of pixels that are likely not recovered correctly from view synthesis \cite{147_qi2018geonet,174_zhou2017unsupervised}, and \cite{155_godard2019digging} used the minimum between forward and backward re-projection error to handle occlusion. 

\subsubsection{Regularization losses} 
\paragraph{Cross-frame consistency} It is applied to fully exploit available connections in data between stereo pairs and sequential frames and improve generalizability by enforcing the network to learn view synthesis in different directions. For example, \cite{162_godard2017unsupervised} performed view synthesis on a view-synthesized image from the stereo's view, aiming to recover the original image from this loop. 
\paragraph{Cross-task consistency} It is applied to regularize the depth prediction by exploiting the correlation with other tasks, e.g., surface normal prediction \cite{161_dharmasiri2017joint}, optical flow prediction \cite{148_yin2018geonet,156_ranjan2019competitive}, and semantic segmentation \cite{160_ramirez2018geometry}. 
\paragraph{Self-regularization} These are loss terms that suppress high-order variations in depth predictions. Edge-aware depth smoothness loss \cite{EdAwSmooth_Heise_2013_ICCV} is one of the most common example \cite{142_kuznietsov2017semi,162_godard2017unsupervised}. They are widespread because, in unsupervised approaches, view-synthesis losses rely on image gradients, which are heavily non-convex and only valid in a local region. In supervised approaches, sparse ground truth leaves a subset of points uncovered. Such a regularization term can smooth out the prediction and broadcast supervision signal to a larger region.

Supervision signals in the literature are mostly from pixel-wise values (e.g., depth/reprojection error) and simple statistics in a local region (e.g., surface normal, SSIM), with heuristic regularization terms addressing the locality of such supervision signals. In contrast, we are introducing a new loss term that is smooth and continuous, overcoming such locality with embedded regularization effect.

\begin{figure*}[t]
\centering
\begin{tabular*}{\textwidth}{@{\extracolsep{\fill} }llllllllllllllll}
  & \footnotesize Image \& LIDAR & & & \footnotesize DORN\cite{dorn_fu2018deep} & & & \footnotesize BTS\cite{bts_lee2019big} & & & \footnotesize Monodepth2\cite{155_godard2019digging} & & & \footnotesize Ours & & \\
\end{tabular*}
    \includegraphics[width=\textwidth]{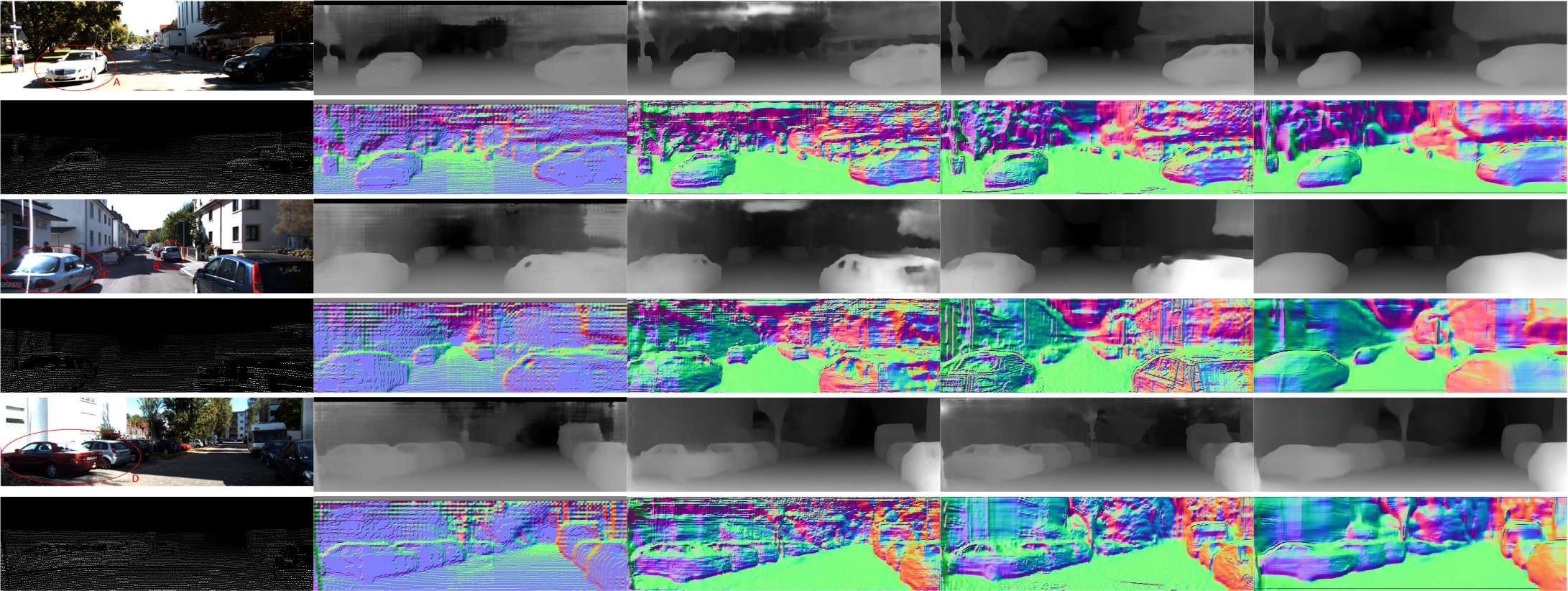}
    \caption{Qualitative result on KITTI dataset. Three samples are shown. Each corresponds to two rows, showing depth prediction and surface normal directions calculated from predicted depth respectively (except the 1st column showing images and LIDAR point-clouds projected on image frame). Regions highlighted in circles, numbered A, B, C, D, are zoomed in with point-cloud view in Fig. \ref{fig:kitti_pcl}.}
    \label{fig:kitti_dep}
\end{figure*}

\section{Proposed Method} \label{sec:method}
Information captured by LIDAR and camera sensors is a discretized sampling of the real environment in points and pixels. The discretization of the two sensors are different, and a common approach of associating them is to project LIDAR points onto the image frame. This approach has two drawbacks. First, it is an approximation to allocate a pixel location for LIDAR points, subject to rounding error and foreground-background mixture error \cite{5_Qiu_2019_CVPR}. Secondly, LIDAR points are much sparser than image pixels, meaning that the supervision signal is propagated from only a small fraction of the image, and surfaces with certain characteristics (e.g., reflective, dark, transparent) are constantly missed due to the limitations of the LIDAR.

To handle the first problem, we evaluate the proposed loss function in the 3D space instead of the image frame. Specifically, we measure the difference between the LIDAR point cloud and the point cloud of image pixels back-projected using the predicted depth. This approach is similar to that of~\cite{154_mahjourian2018unsupervised}, which applied the distance metric of ICP for depth learning. However, since ICP needs the association of point pairs, this approach still suffers from the discretization problem. This problem may not be prominent when both point clouds are from image pixels \cite{154_mahjourian2018unsupervised} but is important when using the sparse LIDAR point cloud. 

We propose to transform the point cloud into a continuous function, and thus the learning problem becomes aligning two functions induced by the LIDAR point cloud and the image depth (point cloud). Our approach alleviates the discretization problem, as shown in Sec. \ref{sec:qualitative} and \ref{sec:quantitative} in more details.  

\subsection{Function construction from a point cloud}
Consider a collection of points, $X=\{(x_i, \ell_X(x_i))\}_{i=1}^n$, with each point $x_i \in \mathbb{R}^3$ and its associated feature vector $\ell_X(x_i) \in \mathcal{I}$, where $(\mathcal{I}, \langle \cdot, \cdot \rangle_{\mathcal{I}})$ is the inner product space of features. 
To construct a function from a point cloud such as $X$, we follow the approach of~\cite{MGhaffari-RSS-19,clark2020nonparametric}. That is
\begin{equation}
    f =  \sum_{i=1}^n \ell_X(x_i) k(\cdot, x_i),
\end{equation}
where $k: \mathbb{R}^3 \times \mathbb{R}^3 \to \mathbb{R}$ is the kernel of a Reproducing Kernel Hilbert Space (RKHS)~\cite{berlinet2004reproducing}.
Then the inner product with function $g$ of point cloud $Z=\{(z_j, \ell_Z(z_j))\}_{j=1}^m$ is given by 
\begin{equation}
    \langle f, g\rangle = \sum_{i=1}^n\sum_{j=1}^m \langle\ell_X(x_i), \ell_Z(z_j)\rangle_{\mathcal{I}} k(x_i, z_j).
\end{equation}
For simplicity, let $c_{ij} := \langle\ell_X(x_i), \ell_Z(z_j)\rangle_{\mathcal{I}}$.
We model the geometric kernel, $k$, using the exponential kernel~\cite[Chapter 4]{rasmussen2006gaussian} as
\begin{equation} \label{eq:exp_kern}
k(x,z) = \sigma\exp\left(\frac{-\lVert x-z\rVert}{s}\right),
\end{equation}
where $\sigma$ and $s$ are tuneable hyperparameters controlling the size and scale, and $\lVert \cdot \rVert$ is the usual Euclidean norm. While there is no specific restrictions on what kernel to use, we found this kernel providing satisfactory result in practice.


\subsection{Continuous 3D loss} \label{sec:constrct_func}
Let $Z$ be the LIDAR point cloud that we use as the ground truth, and $X$ the point cloud from image pixels with depth. We then formulate our continuous 3D loss function as: 
\begin{equation} \label{eq:loss_c3d}
    L_{C3D}(X, Z) = -\sum_{i=1}^n\sum_{j=1}^m c_{ij} k(x_i, z_j),
\end{equation}
i.e. to maximize the inner product. Different from \cite{MGhaffari-RSS-19} which aims to find the optimal transformation in the Lie group to align two functions, we operate on points in $X$. The gradient of $L_{C3D}$ w.r.t. a point $x_i \in X$ is:
\begin{equation}
    \frac{\partial L_{C3D}}{\partial x_i} = -\sum_{j=1}^m ( c_{ij} \frac{\partial k(x_i, z_j)}{\partial x_i} + \frac{\partial c_{ij}}{\partial x_i} k(x_i, z_j) ).
\end{equation}
For the exponential kernel we have: 
\begin{equation}
    \frac{\partial k(x_i, z_j)}{\partial x_i} = k(x_i, z_j) \frac{z_j - x_i}{s \lVert x_i-z_j\rVert} ,
\end{equation}
and for $\frac{\partial c_{ij}}{\partial x_i}$ it depends on the specific form of the inner product of the feature space. 

In our experiments we design two set of features, i.e., $c_{ij} := c_{ij}^v \cdot c_{ij}^n$. The first one is the color in the HSV space denoted as $\ell^v$. We define the inner product in the HSV vector space using the same exponential kernel form and treat $\ell^v(x)$ as a constant. Since the pixel color is invariant w.r.t. its depth, $\frac{\partial c_{ij}^{v}}{\partial x_i} = 0$. 

The second feature is the surface normal, denoted $\ell^n$, and we use a weighted dot product as the inner product of normal features, i.e. 
\begin{equation}
    c_{ij}^{n} := \frac{\ell_X^n(x_i)^\transpose \ell_Z^n(z_j)}{r_X^n(x_i) + r_Z^n(z_j) + \epsilon} ,
\end{equation}
where $\epsilon$ is to avoid numerical instability, and $r^n(x)$ denotes the residual, embedding the smoothness of local surface at $x$, which is further explained in the following.

Given a point $x_i$ with normal vector $l_X^n(x_i)$, the plane defined by the normal is given as: 
\begin{equation}
    N_{x_i} = \{x: x^T l_X^n(x_i) - x_i^T l_X^n(x_i) = 0 \}.
\end{equation}{}
Accordingly, the residual of an arbitrary point $x'$ w. r. t. this local surface is defined as:
\begin{equation}
    r_X^n(x'; x_i) = \frac{\lVert {x'}^T l_X^n(x_i) - x_i^T l_X^n(x_i) \rVert}{\lVert x'-x_i \rVert} \in \left[ 0, 1\right] ,
\end{equation}
which equals to the cosine angle between the line $\overline{x_i x'}$ and the local surface normal. Then the residual of the local surface is defined as:
\begin{equation}
    r_X^n(x_i) = \frac{1}{| U(x_i) |}\sum_{x' \in U(x_i)} r_X^n(x'; x_i)  \in \left[ 0, 1\right] ,
\end{equation}
where $U(x_i)$ is the set of points in the neighborhood of $x_i$, and $|U(x_i)|$ denotes the number of elements in the set (its cardinality). This term equals to the average of the residual using a neighborhood of the local plane.  

The derivative of this kernel w.r.t. the local surface normal vector $\ell_Z^n(x_i)$ is then give by
\begin{equation}
    \frac{\partial c_{ij}^{n}}{\partial \ell_Z^n(x_i)} = \frac{\ell_Z^n(z_j)}{r_X^n(x_i) + r_Z^n(z_j) + \epsilon} .
\end{equation}

From the above analysis, we can see that the continuous 3D loss function produces a gradient that combines position difference and normal-direction differences between ground truth points and predicted points weighted by their closeness in the geometric and the feature space. The proposed method avoids point-to-point correspondences that are not always available in data and provides an inherent regularization that can be adjusted with understandable physical meanings. 


The exponential operations in $L_{C3D}$ result in very large numbers compared with other kinds of losses. For numerical stability, we use logarithm of the 3D loss in practice, i.e.
\begin{equation}
    L_{C3D}'(X, Z) = \log (L_{C3D}(X, Z)).
\end{equation}
The continuous 3D loss can be used for cross-frame supervision, in which case relative camera poses also come into play. For example, we can denote:
\begin{equation}
    L_{C3D_{i,j}}'(X_i, Z_j) = L_{C3D_{i,i}}'(X_i, T_{j}^{i}Z_j).
\end{equation}
where $X_i, Z_i$ denotes point-clouds from camera and from LIDAR at frame $i$, and $T_{j}^{i}\in SE(3)$ transforms points in coordinate $j$ to coordinate $i$. 

\subsection{Network architecture}
To evaluate the effect of the continuous 3D loss function, we modified three state-of-the-art monocular depth prediction approaches: Monodepth2 \cite{155_godard2019digging}, DORN \cite{dorn_fu2018deep}, and BTS \cite{bts_lee2019big}, by simply including the proposed loss function as an extra loss term. DORN and BTS are supervised depth prediction approaches, which are closely related to our work. Monodepth2 is originally an unsupervised approach, which we included to show that our proposed loss is still effective in the presence of photometric losses. For a fair comparison, we added an L1 depth loss to Monodepth2 so that all baselines are supervised.

\begin{figure}
    \centering
    \resizebox{\columnwidth}{!}{
    \begin{tabular*}{\textwidth}{@{\extracolsep{\fill} }llllllllll}
      & \LARGE A & & \LARGE B & & \LARGE C & & & \LARGE D &\\
    \end{tabular*}
    }
    \includegraphics[width=\columnwidth]{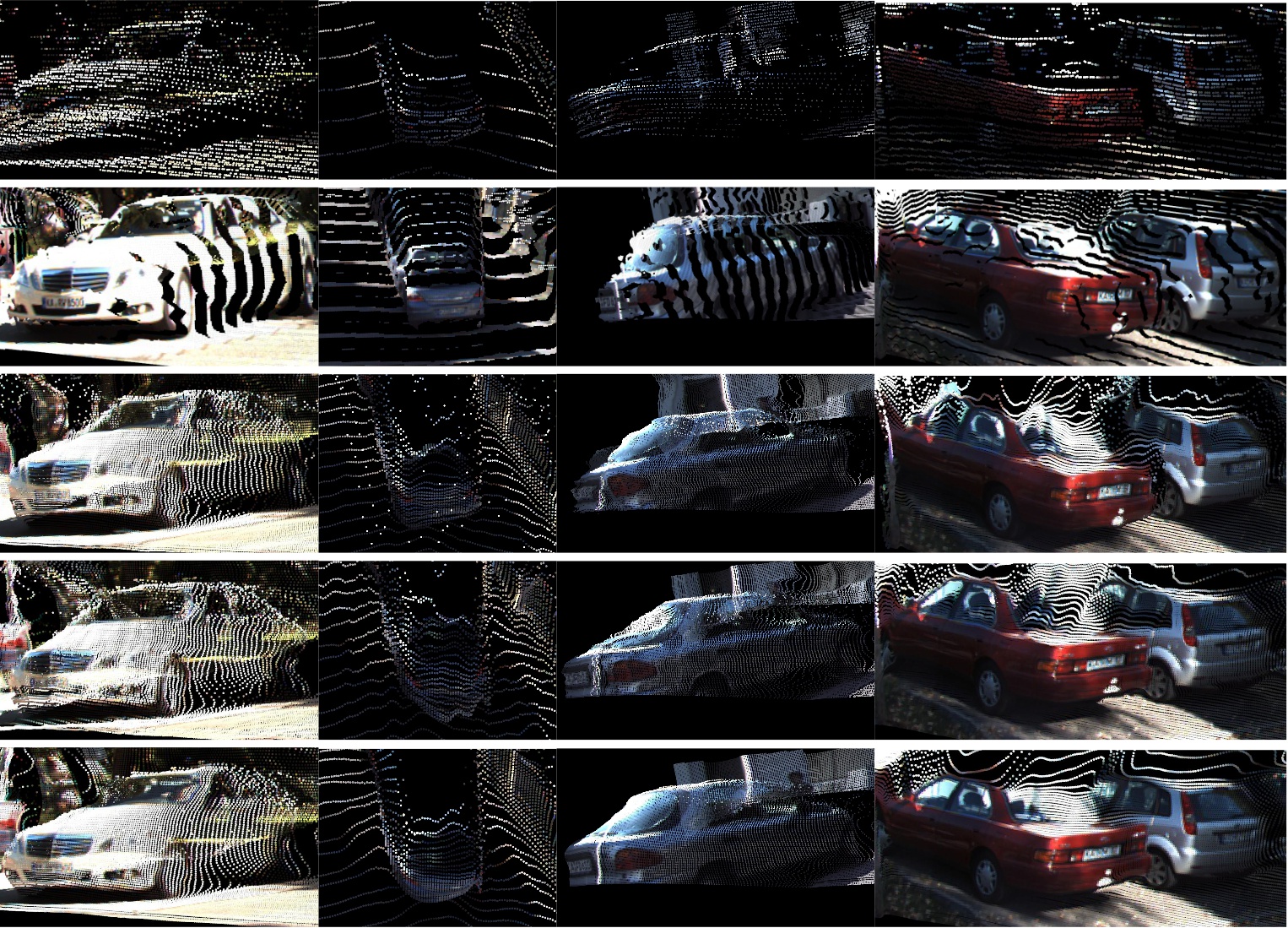}
    \caption{Point-cloud visualization of vehicles circled in Fig. \ref{fig:kitti_dep}. \textbf{From top to bottom}: raw LIDAR colored by image, point clouds generated by DORN~\cite{dorn_fu2018deep}, BTS~\cite{bts_lee2019big}, Monodepth2~\cite{155_godard2019digging}, and \textbf{by our approach}. The four columns correspond to A, B, C, D in Fig. \ref{fig:kitti_dep} respectively. } 
    \label{fig:kitti_pcl}
\end{figure}

\section{Experiments} \label{sec:experiments}
\subsection{Implementation details}
The model is implemented in Pytorch, and the training settings are consistent with the three baseline methods, except that the batch size is set to 3 in all methods. Specifically, the backbone feature extraction networks are ResNet-50\cite{he2016deep}, ResNet-101, and DenseNet-161\cite{huang2017densely}, and the training epochs are 20, 40, and 50 for Monodepth2, DORN, and BTS respectively. 

We implemented a customized operation in Pytorch to efficiently calculate the inner product on GPU, taking advantage of the sparsity of LIDAR point clouds and the double sum. Such computation only induces a small (5\%) time overhead in each iteration. 

In practice, LIDAR point clouds are cropped to only keep the front section in the camera view before calculating $L_{C3D}$. Besides, we can see from \eqref{eq:loss_c3d} that the calculation of the inner product involves a double sum over all point-pairs in two point clouds. To alleviate the computation burden, we discard point-pairs that are far away from each other in image space, of which the geometric kernel value is likely to be very small, hardly contributing to the loss. 

The parameters in $L_{C3D}$ mainly involves $\sigma$ and $s$ in the exponential kernel (\ref{eq:exp_kern}). For the HSV feature kernel, we use $\sigma_v=1$, $s_v=0.2$. For the geometric kernel, we use $\sigma_g=1$, $s_g=s_0d$, where $d$ is the maximum depth in the pair of points involved in the kernel, so that the support of the kernel grows larger for further points. We do not specifically tune the value of $s_0$. Instead we sample it in each iteration of training as $s_0 = 0.01+0.02|\alpha|, \alpha \sim N(0,1)$.

\subsection{KITTI Dataset}
\label{sec:dataset_kitti}
As is common in literature (e.g. \cite{155_godard2019digging, dorn_fu2018deep, bts_lee2019big}), our experiment is conducted using the KITTI dataset \cite{KITTIRaw_Geiger2013IJRR, KITTIDepth}. All three baselines follow Eigen's data split \cite{Eigen_2015_ICCV}, except that Monodepth2 also used Zhou's \cite{174_zhou2017unsupervised} preprocessing to remove static frames in order to avoid degeneration of photometric losses. 

We note that there are two versions of "depth ground truth" in the KITTI dataset. The first is the projection from raw LIDAR point cloud \cite{KITTIRaw_Geiger2013IJRR}, whereas the second one is preprocessed in KITTI 2015 depth benchmark \cite{KITTIDepth}. The latter is denser with fewer errors than raw LIDAR projection. This is due to the accumulation of 11 adjacent LIDAR scans and outlier removal by comparing them with stereo estimations. However, this densified depth ground truth is still semi-dense, i.e., not covering all pixels in images. To highlight our purpose of better leveraging LIDAR point clouds and to make the approach more generalizable to other datasets where such preprocessing is unavailable, we use the raw LIDAR point clouds for training and use the refined and denser depth images for evaluation. This results in 652 images in the test set, which are the frames with refined depth images in Eigen's test split. This setup is different from what is in DORN and BTS; therefore, the baselines' quantitative result is generated by us and not the same as in the original papers. 




\begin{table*}[t]
    \centering
    \captionsetup{justification=centering}
    \caption{The quantitative comparison using Eigen's test split with improved ground truth. \\ $\bullet$ \textbf{Bold} numbers are the best. The rows of ``Improvement'' are w.r.t. the baselines. \\ $\bullet$ The ``Train'' column, ``U'': unsupervised, ``SS'': supervised by stereo disparity, ``LS'': supervised by LIDAR depth, ``DS'': supervised by densified KITTI depth. Our experiments are focused on ``LS'' cases, while results from other supervisions are given for reference. \\ $\bullet$ The ``Source'' column shows where the numbers are from. ``O'': generated by us based on the official open-source implementation. ``U'': generated by us based on unofficial implementation, where we made our best effort to align with the original paper. \\ $\bullet$ Gray results are with the supervision as in the original papers. They are a better reference than the numbers in ``from literature'' section because our experiments are conducted with training setups as similar as possible. In contrast, for example, the numbers of DORN and Monodepth2 in ``from literature'' section are of different backbones from those in our experiments. }
    \footnotesize
    \begin{tabular}{c|l|c|c|cccc|ccc}
        \toprule
        \multicolumn{1}{l}{} & \multicolumn{1}{l}{} & \multicolumn{1}{c}{} & \multicolumn{1}{c}{} & \multicolumn{4}{c}{lower is better} & \multicolumn{3}{c}{higher is better} \\
        \midrule
        \multicolumn{1}{l}{} & Method & Train & Source & Abs Rel & Sq Rel & RMSE & RMSE log & $\delta < 1.25$ & $\delta < 1.25^2$ & $\delta < 1.25^3$ \\
        \midrule
        \multirow{8}{*}{\begin{sideways}from literature\end{sideways}} & DDVO\cite{181_wang2018learning} & U & \cite{155_godard2019digging} & 0.126 & 0.866 & 4.932& 0.185 & 0.851 & 0.958 & 0.986 \\
        & 3net\cite{poggi2018learning} & U & \cite{155_godard2019digging} & 0.102 & 0.675 & 4.293 & 0.159 & 0.881 & 0.969 & 0.991 \\
        & SuperDepth\cite{pillai2019superdepth} & U & \cite{155_godard2019digging} & 0.090 & 0.542 & 3.967 & 0.144 & 0.901 & 0.976 & 0.993 \\
        & Monodepth2\cite{155_godard2019digging} & U & \cite{155_godard2019digging} & 0.090 & 0.545 & 3.942 & 0.137 & 0.914 & 0.983 & 0.995 \\
        & SVSM FT\cite{svsm_luo2018single} & U+SS & \cite{semidepth_amiri2019semi} & 0.077 & 0.392 & 3.569 & 0.127 & 0.919 & 0.983 & 0.995 \\ 
        & semiDepth\cite{semidepth_amiri2019semi} & U+DS & \cite{semidepth_amiri2019semi} & 0.078 & 0.417 & 3.464 & 0.126 & 0.923 & 0.984 & 0.995 \\ 
        & DORN\cite{dorn_fu2018deep} & DS & \cite{semidepth_amiri2019semi} & 0.080 & 0.332 & 2.888 & 0.120 & 0.938 & 0.986 & 0.995 \\ 
        & BTS\cite{bts_lee2019big} & DS & \cite{bts_lee2019big} & \textbf{0.060} & \textbf{0.249} & \textbf{2.798} & \textbf{0.096} & \textbf{0.955} & \textbf{0.993} & \textbf{0.998} \\ 
        \midrule
        
        \midrule
        \multirow{13}{*}{\begin{sideways}from our experiments\end{sideways}} & Monodepth2 (Baseline) & U+LS & O & 0.077 & 0.444 & 3.568 & 0.118 & 0.934 & \textbf{0.988} & \textbf{0.997} \\ 
        & Monodepth2+C3D (Ours)  & U+LS & O & 0.072 & 0.370 & 3.371 & 0.116 & 0.937 & \textbf{0.988} & \textbf{0.997} \\ 
        & \textcolor{gray}{Monodepth2} & \textcolor{gray}{U} & \textcolor{gray}{O} & \textcolor{gray}{0.087} & \textcolor{gray}{0.509} & \textcolor{gray}{3.812} & \textcolor{gray}{0.126} & \textcolor{gray}{0.922} & \textcolor{gray}{0.984} & \textcolor{gray}{0.995} \\
        & \textit{Improvement} & U+LS & O & \textit{6.5\%} & \textit{16.6\%} & \textit{5.5\%} & \textit{1.7\%} & \textit{0.3\%} & \textit{0.0\%} & \textit{0.0\%} \\
        \cmidrule{2-11}
        & BTS (Baseline) & LS & O & 0.071 & 0.342 & 3.341 & \textbf{0.115} & 0.936 & 0.987 & \textbf{0.997} \\
        & BTS+C3D (Ours) & LS & O & \textbf{0.068} & \textbf{0.326} & 3.231 & \textbf{0.115} & \textbf{0.937} & 0.987 & \textbf{0.997} \\ 
        & \textcolor{gray}{BTS}        & \textcolor{gray}{DS} & \textcolor{gray}{O} & \textcolor{gray}{0.063} & \textcolor{gray}{0.268} & \textcolor{gray}{2.896} & \textcolor{gray}{0.101} & \textcolor{gray}{0.949} & \textcolor{gray}{0.991} & \textcolor{gray}{0.998} \\
        & \textit{Improvement} & LS & O & \textit{4.2\%} & \textit{4.7\%} & \textit{3.3\%} & \textit{0.0\%} & \textit{0.1\%} & \textit{0.0\%} & \textit{0.0\%} \\
        
        \cmidrule{2-11}
        
        & DORN (Baseline)  & LS & U & 0.127 & 0.474 & 3.420 & 0.153 & 0.900 & 0.985 & 0.996 \\
        & DORN+C3D (Ours) & LS & U & 0.117 & 0.409 & \textbf{3.155} & 0.142 & 0.916 & \textbf{0.988} & \textbf{0.997} \\
        & \textcolor{gray}{DORN} & \textcolor{gray}{DS} & \textcolor{gray}{U} & \textcolor{gray}{0.110} & \textcolor{gray}{0.358} & \textcolor{gray}{3.064} & \textcolor{gray}{0.133} & \textcolor{gray}{0.927} & \textcolor{gray}{0.991} & \textcolor{gray}{0.998} \\
        & \textit{Improvement} & LS & U & \textit{7.9\%} & \textit{13.7\%} & \textit{7.8\%} & \textit{7.2\%} & \textit{1.8\%} & \textit{0.3\%} & \textit{0.1\%} \\
       
        \bottomrule
    \end{tabular}
    \label{tab:kitti}
    \squeezeup
\end{table*}

\subsection{Quantitative results and analysis} \label{sec:quantitative}
Consistent with the literature \cite{162_godard2017unsupervised}, depth is truncated at $80\m$ maximum. We also crop a portion of the images as done in \cite{144_garg2016unsupervised} before evaluation. The same setup can also be found in Monodepth2, DORN, and BTS. The definition of all metrics is consistent with those of \cite{140_eigen2014depth}.
In Table~\ref{tab:kitti}, the quantitative comparison of our method with the baselines and other state-of-the-art approaches is reported. Improvement is achieved by simply adding our proposed continuous 3D loss function to all three baseline methods. 
\begin{remark}
We note that our approach does not outperform BTS results reported in the literature, as shown in Table~\ref{tab:kitti}. The reason is that BTS is trained using refined and densified KITTI depth. When supervised by raw LIDAR depth, Our experiment shows that the proposed method can improve BTS, DORN, and Monodepth2. Our accuracy lies between baselines trained using raw LIDAR depth and those trained using densified depth. It implies that the ideal case is to have dense supervision. Our method acts as a surrogate to dense supervision when we only have access to sparse supervision. 
\end{remark}

\begin{table*}[t]
    \centering
    \caption{Quantitative comparison for ablation study on the effect of surface normal kernel.}
    \footnotesize
    \begin{tabular}{l|l|cccc|ccc}
        \toprule
        \multicolumn{2}{l}{} & \multicolumn{4}{c}{lower is better} & \multicolumn{3}{c}{higher is better} \\
        \midrule
        Dataset & Method & Abs Rel & Sq Rel & RMSE & RMSE log & $\delta < 1.25$ & $\delta < 1.25^2$ & $\delta < 1.25^3$ \\
        \midrule
        \multirow{3}{*}{KITTI} & Monodepth2 (Baseline) & 0.077 & 0.444 & 3.568 & 0.118 & 0.934 & \textbf{0.988} & \textbf{0.997} \\ 
        & $L_{C3D}^{v}$   & 0.075 & 0.404 & 3.481 & 0.117 & 0.935 & \textbf{0.988} & \textbf{0.997} \\
        & $L_{C3D}^{nv}$ & \textbf{0.072} & \textbf{0.370} & \textbf{3.371} & \textbf{0.116} & \textbf{0.937} & \textbf{0.988} & \textbf{0.997} \\
        \bottomrule
    \end{tabular}
    \label{tab:abla_kitti}
    \squeezeup
\end{table*}
\subsection{Qualitative results and analysis} \label{sec:qualitative}
In order to show the effect of the new continuous 3D loss intuitively, in Fig. \ref{fig:kitti_dep} we listed a few samples from the KITTI dataset. Each sample includes the RGB image, the raw LIDAR scan, and the predicted depth and corresponding surface normal directions from the baselines and our method. We only show our results based on Monodepth2 network and omit our result based on DORN and BTS due to page limit.

\subsubsection{Depth view}
We observe that both Monodepth2 and BTS predict incorrectly at the vehicle-window area from the depth prediction images. It creates ``holes'' in the depth map and fails to recover the full object contours, as in the second and third examples. This area is not handled well by previous methods because 
\begin{itemize}
    \item The window area is a non-Lambertian surface with an inconsistent appearance at different viewing angles; therefore, photometric losses do not work.
    \item LIDAR does not receive a good reflection from glasses, as can be seen from Fig. \ref{fig:kitti_dep}; therefore, no supervision from the ground truth is available. 
    \item The window area's color is usually not consistent with other parts of the vehicle body, further failing appearance-based depth smoothness terms.
\end{itemize}{} 
In contrast, our continuous 3D loss function provides supervision from all nearby points, thus overcoming the problem and providing inherent smoothness. The window-area is predicted correctly with full object shape preserved from our predictions.

DORN presents fewer "holes" and irregular contours in the depth images than the other two baselines. Its classification formulation restricts the possible distortions. It has a side effect that the predicted depths are all from a predefined discrete set of values, which can be observed more intuitively in surface-normal view in Sec. \ref{sec:surface_normal_view} and point-cloud view in Sec. \ref{sec:pointcloud_view}. 

\subsubsection{Surface-normal view}\label{sec:surface_normal_view}
The surface-normal view provides a better visualization of 3D structures and local smoothness. The second row of each sample in Fig. \ref{fig:kitti_dep} shows the surface normal direction calculated from the predicted depth. Despite the existence of regularizing smoothness term, the baseline method, Monodepth2, still produces many textures inherited from the color space. This is because the edge-aware smoothness loss is down-weighted at high-gradient pixels. BTS shows less, but still visible, texture and artifacts from color space in the normal map, and the inconsistency in window-area is apparent in surface-normal view. In contrast, our method does not produce such textures while still preserving the 3D structures with a clear shift between different surfaces. DORN produces almost-uniform normal images because the predicted depth value range is discrete.

\subsubsection{Point-cloud view}\label{sec:pointcloud_view}
By back-projecting image pixels using predicted depth to 3D space, we can recover the scene's point cloud. This view allows us to inspect how well the depth prediction recovers the 3D geometry in the real world. This is important as the pixel-clouds could provide a denser alternative to accurate-but-sparse LIDAR point clouds, benefiting 3D object detection, as indicated in \cite{21_Wang_2019_CVPR}. 

Fig. \ref{fig:kitti_pcl} shows four examples. They cover both near and far objects and cases of over-exposure and color-blend-in with the background. We can see that the raw LIDAR scans are quite sparse on dark vehicle bodies and glass surfaces, posing challenges on using such data as ground truth for depth learning. ``Holes'' in predicted depth map transform to unregulated noise points in 3D view. Compared with the Monodepth2 and BTS baselines, our method produces point clouds with higher quality in both glass and non-glass areas, with a smooth surface and geometric structure consistent with the real vehicles. The shape distortion of DORN-produced point-clouds is also small, but the points all lie on some common vertical planes due to the discrete depth prediction, making the point-clouds unrealistic. Our method is capable of producing well-shaped point-clouds while not bound to this limitation.

\subsubsection{Summary}
In the qualitative comparison, we show that our method predicts depth with a smoother shape and less distortion, especially in reflective and transparent areas. This improvement is not fully presented in the quantitative analysis because the ground truth depth in those areas is generally missing. 


\begin{figure}[b]
\begin{tabular*}{\columnwidth}{l@{\extracolsep{\fill}}l}
  \cell{{\rotatebox[origin=c]{90}{\parbox[c]{25pt}{\scriptsize Image \& LIDAR}}}    \\ 
        {} \\
        {\rotatebox{90}{\tiny S-Monodepth2}}  \\
        {} \\
        {\rotatebox{90}{\scriptsize $L_{C3D}^v$}} \\
        {} \\
        {\rotatebox{90}{\scriptsize $L_{C3D}^{nv}$}}}           
    & \cell{\includegraphics[width=0.95\columnwidth]{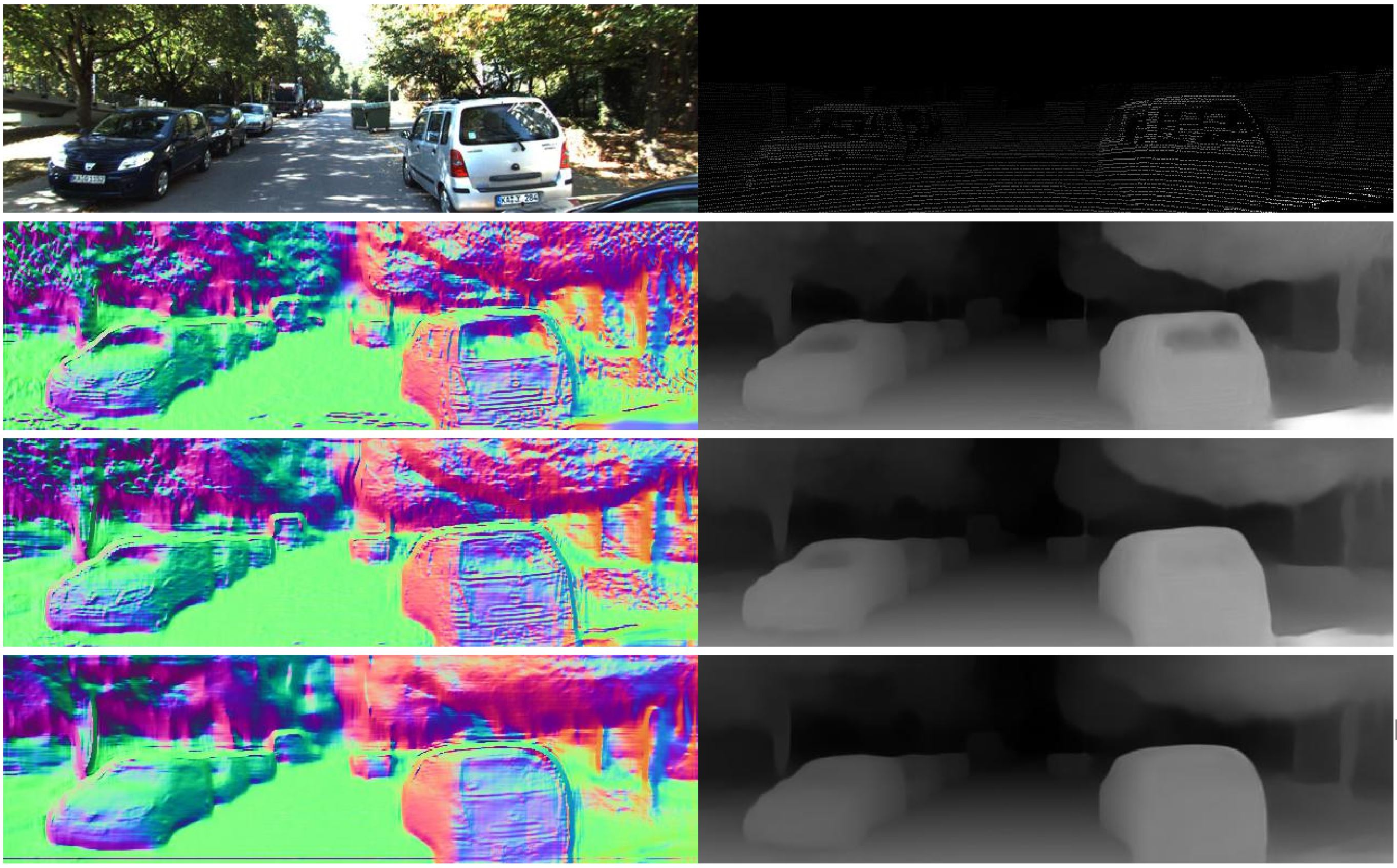}} 
\end{tabular*}
    \caption{Visualization of the effect of surface normal kernel. Except for the 1st column, left images are surface normals, and on the right side, corresponding predicted depth images are shown.}
    \label{fig:ablation}
\end{figure}

\subsection{Ablation study}
We now take a closer look at different configurations in the function constructed from point clouds. Here we mainly investigate the effect of the surface normal kernel. We denote the continuous 3D loss without surface normal kernel as
\begin{equation} \label{eq:loss_c3d_base}
    L_{C3D}^{v}(X, Z) = -\sum_{i=1}^n\sum_{j=1}^m c_{ij}^v k(x_i, z_j)
\end{equation}{}
and the one with surface normal kernel as:
\begin{equation} \label{eq:loss_c3d_with_nkern}
    L_{C3D}^{nv}(X, Z) = -\sum_{i=1}^n\sum_{j=1}^m c_{ij}^n c_{ij}^v k(x_i, z_j)
\end{equation}{}
The quantitative comparison is in Table~\ref{tab:abla_kitti}, following the same setup as Fig. \ref{tab:kitti} and a data sample is shown in Fig. \ref{fig:ablation} for visualization. We only show results using the Monodepth2 baseline due to space limitation. While the continuous loss $L_{C3D}^{v}$ improved upon the baseline by exploiting the correlation among points, the prediction still produces artifacts caused by textures in color space. The surface normal kernel is sensitive to local noises and distinguishes between different parts of the 3D geometry, producing more geometrically plausible predictions.

\section{Conclusion} \label{sec:conclusions}
We proposed a new continuous 3D loss function for monocular single-view depth prediction. The proposed loss function addresses the gap between dense image prediction and sparse LIDAR supervision. We achieved this by transforming point clouds into continuous functions and aligning them via the function space's inner product structure. By simply adding this new loss function to existing network architectures, the accuracy and geometric consistency of depth predictions are improved significantly on all three state-of-the-art baseline networks that we tested. The evaluation shows that our contribution is orthogonal to the progress in depth prediction network designs and that our work can benefit general depth prediction networks by applying the continuous 3D loss as a plug-in module. 

Future work includes representation learning for features used in the proposed loss function to bring further improvements. Finally, exploring the benefits of the improved depth prediction for 3D object detection is another interesting research direction.

\section*{ACKNOWLEDGMENT}
This article solely reflects the opinions and conclusions of its authors and not TRI or any other Toyota entity.

\bibliographystyle{bib/IEEEtran}
\IEEEtriggeratref{32}
\bibliography{bib/strings-abrv,bib/ieee-abrv,bib/refs_depth}

\end{document}